\documentclass[runningheads]{llncs}
\usepackage[T1]{fontenc}
\usepackage{graphicx,verbatim}
\usepackage{hyperref}
\usepackage{amssymb}
\usepackage{algorithm}
\usepackage{algpseudocode}
\usepackage{subcaption}
\usepackage{graphicx}
\usepackage{amsmath}
\usepackage{multirow}
\usepackage{bbding}
\usepackage{svg}
\usepackage{booktabs}
\usepackage{float}

\begin{document}

\title{ACCURATE: Arbitrary-shaped Continuum Reconstruction Under Robust Adaptive Two-view Estimation}

\titlerunning{ACCURATE: Robust Two-View Reconstruction of Continuum Structures}
%

\author{
Yaozhi Zhang\inst{1} \and
Shun Yu\inst{1} \and
Yugang Zhang\inst{1} \and
Yang Liu\inst{1}
}

\authorrunning{Yaozhi Zhang et al.}

\institute{
Global Institute of Future Technology, 
Shanghai Jiao Tong University, Shanghai, China\\
\email{\{zhangyaozhi, shunyu2024, zyg202307, yang.liu1\}@sjtu.edu.cn}
}
\maketitle              

\begin{abstract}

Accurate reconstruction of arbitrary-shaped long slender continuum bodies, such as guidewires, catheters and other soft continuum manipulators, is essential for accurate mechanical simulation. However, existing image-based reconstruction approaches often suffer from limited accuracy because they often underutilize camera geometry, or lack generality as they rely on rigid geometric assumptions that may fail for continuum robots with complex and highly deformable shapes. To address these limitations, we propose ACCURATE, a 3D reconstruction framework integrating an image segmentation neural network with a geometry-constrained topology traversal and dynamic programming algorithm that enforces global biplanar geometric consistency, minimizes the cumulative point-to-epipolar-line distance, and remains robust to occlusions and epipolar ambiguities cases caused by noise and discretization. Our method achieves high reconstruction accuracy on both simulated and real phantom datasets acquired using a clinical X-ray C-arm system, with mean absolute errors below 1.0 mm. An anonymized repository containing the code and datasets is available at \url{https://anonymous.4open.science/r/ACCURATE}.


\end{abstract}
\section{Introduction}

Accurate 3D reconstruction of arbitrary-shaped long slender continuum bodies plays a fundamental role in computer assisted intervention and other soft robotics applications~\cite{diezinger20223d,robertshaw2025world}. Existing reconstruction strategies can be broadly categorized into sensor-based and imaging-based approaches~\cite{sahu2021shape}. Sensor-based approaches reconstruct the shape by measuring distributed strain or pose along the device, thus achieve excellent accuracy~\cite{ourak2021fusion,lu2023robust,ha2023sensor} but are hard to generalize to different types of continuum bodies, e.g., in neuro-interventional procedures~\cite{harrigan2012handbook}. Consequently, shape reconstruction in these settings must primarily rely on imaging-based approaches.

However, existing imaging-based approaches still face notable limitations. Learning-based methods handle multi-view information implicitly, treating camera geometry as a soft prior—either injected via token embeddings or imposed through loss functions rather than enforcing it as a hard constraint derived from the imaging model~\cite{miyato2023gta,cheng2025monster,wang2025vggt,yang2025fast3r,yao2018mvsnet,altingovde20223d}, may lead to suboptimal reconstruction accuracy. While traditional epipolar methods explicitly enforce geometric constraints and are tailored for reconstructing long, slender continuum bodies~\cite{diezinger20223d,baert2003three,hoffmann2012semi,hoffmann2015electrophysiology}, they depend on simplified imaging assumptions, e.g., known point order and precise camera calibration ensuring intersection between epipolar line and point, or only concern intersection points but not global epipolar-point distance information~\cite{ambrosini20153d,hoffmann2013reconstruction}. Such rigid formulations and assumptions can break down in the presence of large deformations, occlusions, or calibration inaccuracies, as strict point-level correspondence enforcement may overlook the global structural correspondence between two views~\cite{lu2021toward}.

To overcome these limitations, we propose \textbf{ACCURATE}, a biplanar imaging-based 3D reconstruction framework for arbitrary shaped long slender continuum bodies. Unlike end-to-end learning methods, ACCURATE comprises a segmentation module and a segmentation-driven reconstruction module, jointly designed to integrate structural understanding with explicit geometric enforcement. The segmentation module, built upon Topology-aware Segmentation Network (TSN), extracts centerlines of arbitrary shaped long slender continuum bodies across diverse backgrounds. The reconstruction module performs Geometry-consistent Curve Topology Traversal (GCTT) to infer topological ordering from unordered masks, followed by Epipolar-constrained Dynamic Programming (ECDP), which enforces strict biplanar geometric consistency by globally minimizing point-to-epipolar-line deviations. Through this unified design, ACCURATE simultaneously preserves global structural coherence and adheres to camera geometry, enabling accurate and robust 3D reconstruction even under occlusions, missing intersections, and degenerate epipolar configurations where conventional local point-to-point correspondence methods often fail~\cite{cafaro2024two}.

Experiments on both simulated data and real phantom data with accurate 3D ground truth demonstrate the improved accuracy and generalization of ACCURATE across diverse continuum shapes. The real phantom dataset is acquired using a calibrated GE C-arm system, enabling precise 3D spatial annotations. Notably, existing datasets for long slender continuum bodies (e.g., vascular guidewires) are often private or lack camera parameters and 3D ground truth~\cite{jianu2024guide3d}. To bridge this gap, we publicly release our simulated and real phantom datasets, providing biplane X-ray images together with calibrated camera parameters and accurate 3D annotations.

Our contributions are summarized as follows:
\begin{enumerate}
    \item We propose \textbf{ACCURATE}, a decoupled perception–geometry reconstruction framework for arbitrary-shaped long slender continuum structures across diverse backgrounds.
    
    \item We propose a global correspondence formulation solved via dynamic programming that integrates hard geometric constraints with a sub-pixel refinement operator. This approach enforces global geometric consistency, enabling robust reconstruction under occlusions, discretization artifacts, and ambiguous epipolar intersections.
    
    \item We release a publicly available code and dataset for slender continuum reconstruction, providing calibrated multi-view images and accurate 3D annotations to support standardized evaluation and fair comparison.
\end{enumerate}

\section{Method}

The proposed ACCURATE decomposes the reconstruction task into three sequential stages: 
(i) Topology-aware Segmentation Network (TSN), 
(ii) Geometry-consistent Curve Topology Traversal (GCTT), and 
(iii) Epipolar-constrained Dynamic Programming (ECDP).

\begin{figure}
    \centering
    \includegraphics[width=0.98\linewidth]{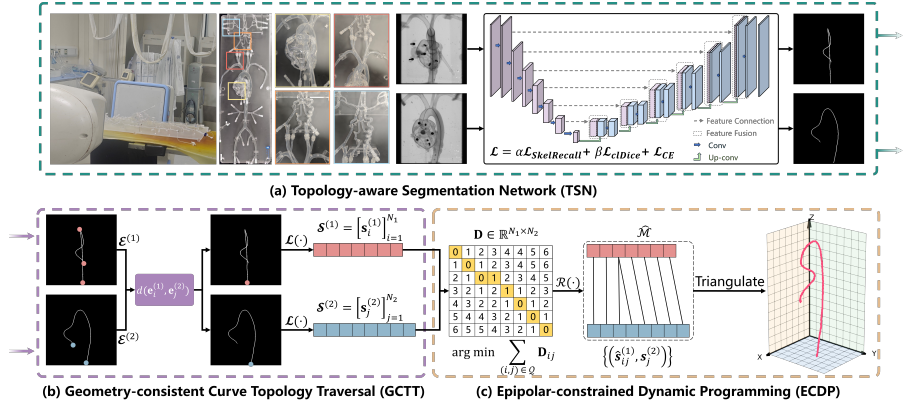}
    \caption{Module (a) takes the original images as input and extracts centerline masks. Module (b) converts the centerlines into topology-ordered point sequences. Module (c) establishes cross-view point correspondences to reconstruct the 3D shape.}
    \label{fig:overview}
\end{figure}

\subsection{Topology-aware Segmentation Network (TSN)}
To address the fragmentation issues often observed in slender guidewire segmentation, we employ an 8-stage nnUNet~\cite{isensee2021nnu} as our backbone (Fig.~\ref{fig:overview}(a)). We optimize a composite objective function, denoted as $\mathcal{L}_{Total}$, to explicitly enforce structural connectivity:

\begin{equation}
\mathcal{L}_{Total} =  \alpha \mathcal{L}_{SkelRecall} + \beta \mathcal{L}_{clDice} + \mathcal{L}_{\text{CE}} 
\label{eq:topo_loss}
\end{equation}

\noindent where $\mathcal{L}_{\text{CE}}$ is the standard cross-entropy loss. To prevent topological fractures, $\mathcal{L}_{SkelRecall}$~\cite{kirchhoff2024skeleton} enforces local integrity by maximizing the overlap between the prediction $\mathbf{P}$ and a pre-computed dilated ``tubed'' skeleton $\mathbf{S}^{tubed}$:

\begin{equation}
\mathcal{L}_{SkelRecall} = 1 - \frac{\sum \mathbf{P} \mathbf{S}^{tubed}}{\sum \mathbf{S}^{tubed} + \epsilon}
\end{equation}

\noindent Complementarily, $\mathcal{L}_{clDice}$~\cite{shit2021cldice} aligns global topology by minimizing the divergence between the soft-skeleton of the prediction and the ground truth skeleton ($\mathbf{S}_{\mathbf{L}}$):

\begin{equation}
\mathcal{L}_{clDice} = 1 - \frac{2 T_{prec}(\mathbf{S}_{\mathbf{P}}, \mathbf{V}_{\mathbf{L}}) T_{sens}(\mathbf{S}_{\mathbf{L}}, \mathbf{V}_{\mathbf{P}})}{T_{prec}(\mathbf{S}_{\mathbf{P}}, \mathbf{V}_{\mathbf{L}}) + T_{sens}(\mathbf{S}_{\mathbf{L}}, \mathbf{V}_{\mathbf{P}})}
\end{equation}

\noindent where $T_{prec}$ and $T_{sens}$ denote topology precision and sensitivity, respectively. We set the weights $\alpha=\beta=1.0$.
Finally, the predicted probability maps are binarized and skeletonized to one-pixel-wide centerlines, ensuring geometrically consistent structures for the topology traversal in Sec.~\ref{sec:GCTT}.

\subsection{Geometry-consistent Curve Topology Traversal (GCTT)}
\label{sec:GCTT}
\begin{figure}
    \centering
    \includegraphics[width=0.98\linewidth]{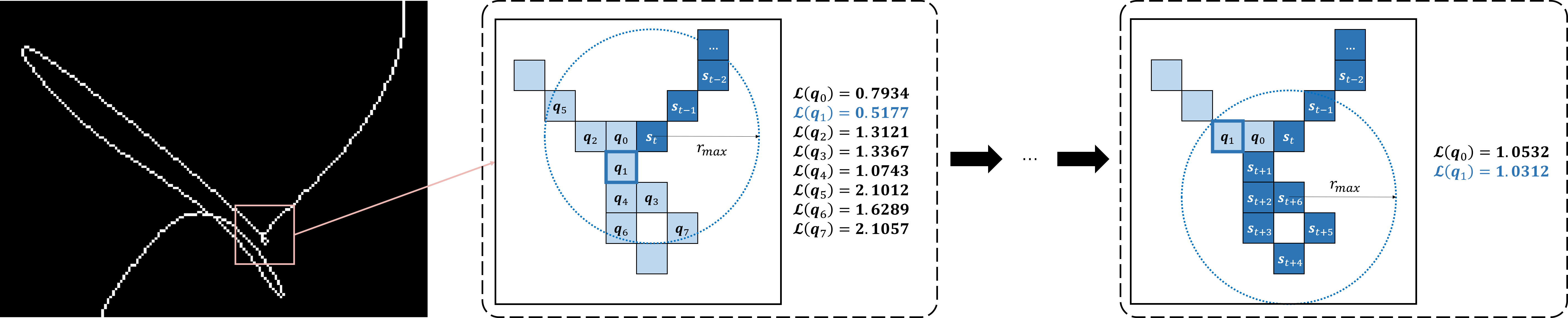}
    \caption{Illustration of the GCTT point selection strategy. From the current points $\mathcal{W}_t=\{\mathbf{s}_{t-m+1},\dots,\mathbf{s}_t\}$, multiple candidate successors may exist. Geometric loss $\mathcal{L}(\cdot)$ ensures smooth and consistent curve traversal.}
    \label{fig:GCTT}
\end{figure}

To achieve precise 3D reconstruction of long slender continuum bodies, strict epipolar correspondence between the two image planes is required. However, an epipolar line in one view may intersect the projected curve multiple or zero times in the other view due to intricate configurations such as self-occlusions, looping topologies, or geometric discrepancies caused by calibration uncertainties, leading to ambiguous local associations~\cite{zhang1998determining}. Recovering the intrinsic topological order of the curve in each view provides a global structural prior that enables consistent correspondence in the subsequent optimization stage. Building on this motivation, we present GCTT. Given segmentation masks $\mathbf{M}^{(k)} \in \{0,1\}^{H\times W}$ from two views $k\in\{1,2\}$ predicted by TSN, we extract unordered point sets $\mathcal{P}^{(k)}=\{\mathbf{p}_n^{(k)}\}_{n=1}^{N_k}$. The endpoint sets are $\mathcal{N}^{(k)}=\{\mathbf{n}^{(k)} \in \mathcal{P}^{(k)} \mid \text{deg}(\mathbf{n}^{(k)})=1\}$. We measure cross-view geometric consistency using the point-to-epipolar-line distance $d(\mathbf{n}_i^{(1)}, \mathbf{n}_j^{(2)})
= \frac{\left| (\tilde{\mathbf{n}}_i^{(1)})^\top \mathbf{F} \tilde{\mathbf{n}}_j^{(2)} \right|}
{\sqrt{\left(\mathbf{F} \tilde{\mathbf{n}}_j^{(2)}\right)_1^2 + \left(\mathbf{F} \tilde{\mathbf{n}}_j^{(2)}\right)_2^2}}$, where $\tilde{\mathbf{n}}=(u,v,1)^\top$ denotes the homogeneous coordinate of $\mathbf{n}$, and fundamental matrix $\mathbf{F}=\mathbf{K}_1^{-\top}[\mathbf{t}]_{\times}\mathbf{R}\mathbf{K}_2^{-1}$. The selected endpoint pair $(\hat{i},\hat{j})=\arg\min_{\mathbf{n}_i^{(1)}\in\mathcal{N}^{(1)},\mathbf{n}_j^{(2)}\in\mathcal{N}^{(2)}} d(\mathbf{n}_i^{(1)},\mathbf{n}_j^{(2)})$. Given the identified starting $\mathbf{s}_0\in\{\mathbf{n}_{\hat{i}}^{(1)}, \mathbf{n}_{\hat{j}}^{(2)}\}$, we recover the ordered topological sequence by iteratively expanding from the current point. Let $\mathcal{V}$ denote the set of visited pixels, and let $\mathbf{s}_t$ be the current point at step $t$. The candidate set $\mathcal{C}(\mathbf{s}_t)$ is defined as all unvisited pixels lying within an annular neighborhood $\mathcal{C}(\mathbf{s}_t)=\{\mathbf{q}\in\mathcal{P}\setminus\mathcal{V}\mid \|\mathbf{q}-\mathbf{s}_t\|\leq r_{\max}\}$,
where larger $r_{\max}$ increases the perceptual coverage and helps bypass occluded or missing points but also introduces higher computational cost.
If $|\mathcal{C}(\mathbf{s}_t)|=1$, the unique candidate is selected. If multiple candidates exist, we choose the next point by minimizing a composite geometric loss $\mathbf{s}_{t+1}=\arg\min_{\mathbf{q}\in\mathcal{C}(\mathbf{s}_t)}\mathcal{L}(\mathbf{q})$, where the loss is
\begin{equation}
    \mathcal{L}(\mathbf{q})=\kappa(\mathcal{W}_t\cup\mathbf{q})+\lambda_a\mathcal{A}(\mathbf{s}_{t-1},\mathbf{s}_t,\mathbf{q})+\lambda_d\|\mathbf{q}-\mathbf{s}_t\|.
\end{equation}
Here $\mathcal{W}_t=\{\mathbf{s}_{t-m+1},\dots,\mathbf{s}_t\}$ denotes a sliding window of the most recent $m$ points. The curvature term $\kappa(\mathcal{W}_t\cup\mathbf{q})$, which measures curvature of the best-fit quadratic curve, enforces local smoothness and the angle penalty is defined as $\mathcal{A}(\mathbf{s}_{t-1},\mathbf{s}_t,\mathbf{q})=1-((\mathbf{s}_t-\mathbf{s}_{t-1})^\top(\mathbf{q}-\mathbf{s}_t))/(\|\mathbf{s}_t-\mathbf{s}_{t-1}\|\,\|\mathbf{q}-\mathbf{s}_t\|)$, which discourages abrupt direction changes. The last term penalizes excessively long jumps between successive points. $\lambda_a$ and $\lambda_d$ are fixed weighting parameters (we choose $\lambda_a=1.0$, $\lambda_d=0.5$ and $r_{\max}=50$), resulting in an ordered topological sequence $\mathcal{S}=[\mathbf{s}_0,\mathbf{s}_1,\dots,\mathbf{s}_T]$. The process and loss are shown in Fig. \ref{fig:GCTT}.

\subsection{Epipolar-constrained Dynamic Programming (ECDP)}
After GCTT, we obtain two topology-consistent ordered pixel sequences $\mathcal{S}^{(1)} = [\mathbf{s}_i^{(1)}]_{i=1}^{N_1}$ and $\mathcal{S}^{(2)} = [\mathbf{s}_j^{(2)}]_{j=1}^{N_2}$. We construct a cost matrix $\mathbf{D}\in\mathbb{R}^{N_1\times N_2}$, where $\mathbf{D}_{ij}=d(\mathbf{s}_i^{(1)},\mathbf{s}_j^{(2)})$ and ECDP computes a globally optimal order-preserving correspondence by dynamic programming. Element $\mathbf{C}_{ij}$ in accumulated cost matrix $\mathbf{C}\in\mathbb{R}^{N_1\times N_2}$ denotes the minimum cumulative cost of matching the prefixes $[\mathbf{s}_1^{(1)},\dots,\mathbf{s}_i^{(1)}]$ and $[\mathbf{s}_1^{(2)},\dots,\mathbf{s}_j^{(2)}]$. The boundary conditions are initialized by cumulative summation along the first row and first column. For each interior entry, the recurrence is defined as $\mathbf{C}_{ij}=\mathbf{D}_{ij}+\min(\mathbf{C}_{i-1,j},\mathbf{C}_{i,j-1 },\mathbf{C}_{i-1,j-1})$, equivalent to solving $\mathcal{M}=\arg\min_{\mathcal{Q}}\sum_{(i,j)\in\mathcal{Q}}\mathbf{D}_{ij},\text{s.t. } \mathcal{Q} \text{ is monotone in } (i,j)$, introduced by Baur \textit{et al.}~\cite{baur2016automatic} while our dynamic programming formulation provides a more efficient solution. However, due to occlusions, discretization, or local segmentation irregularities, the dynamic programming path may contain horizontal or vertical segments, leading to one-to-many or many-to-one correspondences, as also discussed in~\cite{baur2016automatic}. To enforce a unique correspondence for triangulation, we refine such degenerate matches by local interpolation. For a horizontal DP segment 
$\mathcal{O}=\{(i^*,j_1),\dots,(i^*,j_2)\}$, 
we define a distance-weighted linear refinement operator 
$\mathcal{R}(\cdot)$ to convert one-to-many correspondences 
into unique continuous matches. 
The refined view 1 correspondence $\hat{\mathbf{s}}_{i^*j}^{(1)}$ is derived by incorporating the geometric constraints from adjacent rows $i^* \pm1$: $\hat{\mathbf{s}}_{i^*j}^{(1)} 
= \mathcal{R}\!\left(
\mathbf{s}_{i^*-1}^{(1)},\mathbf{s}_{i^*}^{(1)},\mathbf{s}_{i^*+1}^{(1)};\,
\mathbf{D}_{i^*-1,j},\mathbf{D}_{i^*j},\mathbf{D}_{i^*+1,j}
\right),j\in[j_1,j_2]$.
$\mathcal{R}(\cdot)$ performs a weighted interpolation anchored at index $j^* = \arg\min_{j\in[j_1,j_2]}\mathbf{D}_{i^*j}$, defined formally as: 
\begin{equation}
\mathcal{R}(\cdot)=
\begin{cases}
\dfrac{\mathbf{D}_{i^*j}\mathbf{s}_{i^*-1}^{(1)}
+\mathbf{D}_{i^*-1,j}\mathbf{s}_{i^*}^{(1)}}
{\mathbf{D}_{i^*-1,j}+\mathbf{D}_{i^*j}}, 
& j<j^*,\\

\mathbf{s}_{i^*}^{(1)}, 
& j=j^*,\\

\dfrac{\mathbf{D}_{i^*+1,j}\mathbf{s}_{i^*}^{(1)}
+\mathbf{D}_{i^*j}\mathbf{s}_{i^*+1}^{(1)}}
{\mathbf{D}_{i^*j}+\mathbf{D}_{i^*+1,j}}, 
& j>j^*,
\end{cases}
\end{equation}

\begin{figure}[h]
    \centering
    \begin{subfigure}[b]{0.54\linewidth}
        \centering
        \includegraphics[width=\linewidth]{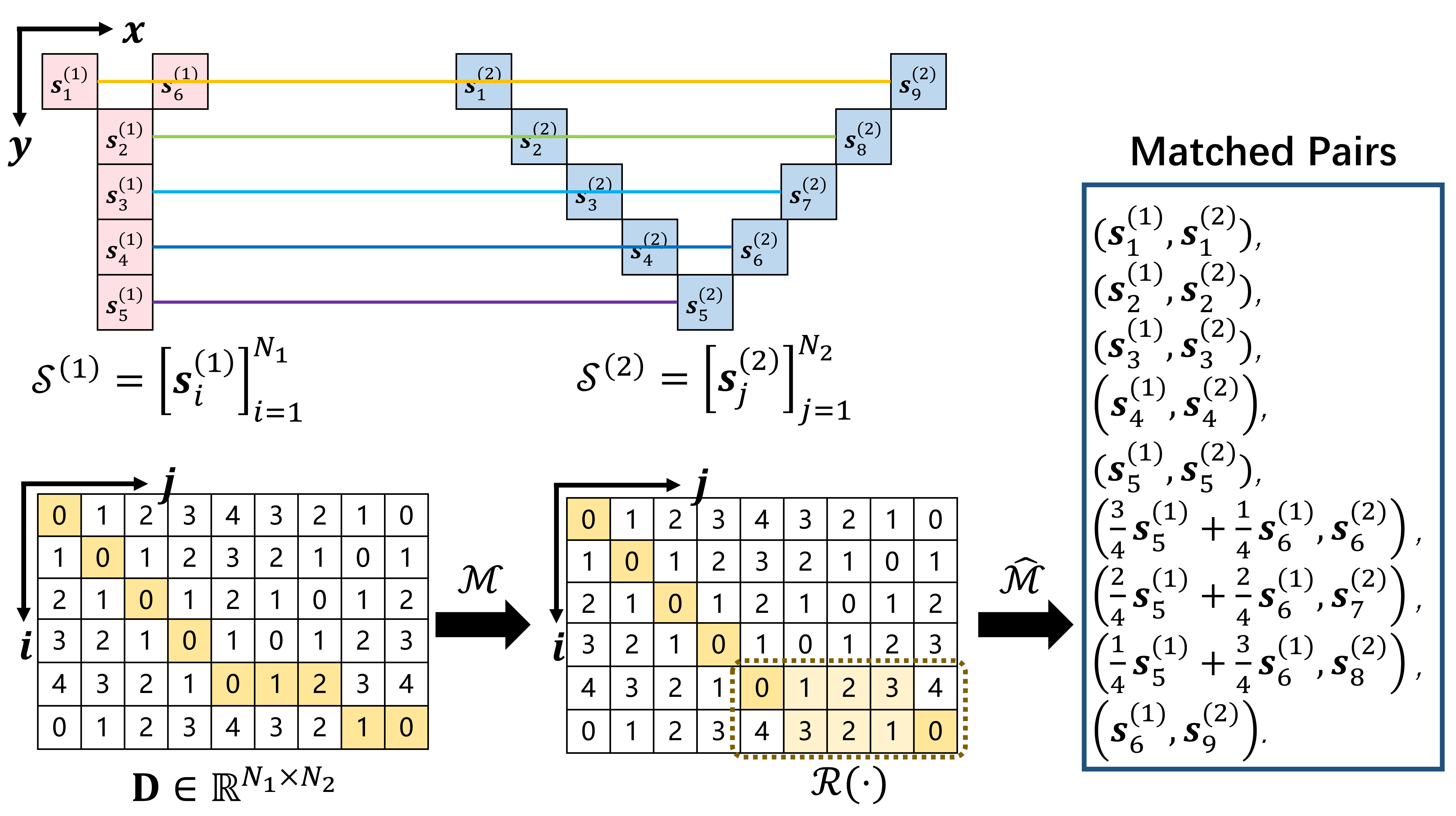}
        \caption{An example of occlusion in view 1. Missing points are constructed by neighbor points in view 1.}
        \label{fig:ECDP_1}
    \end{subfigure}
    \hfill
    \begin{subfigure}[b]{0.44\linewidth}
        \centering
        \includegraphics[width=\linewidth]{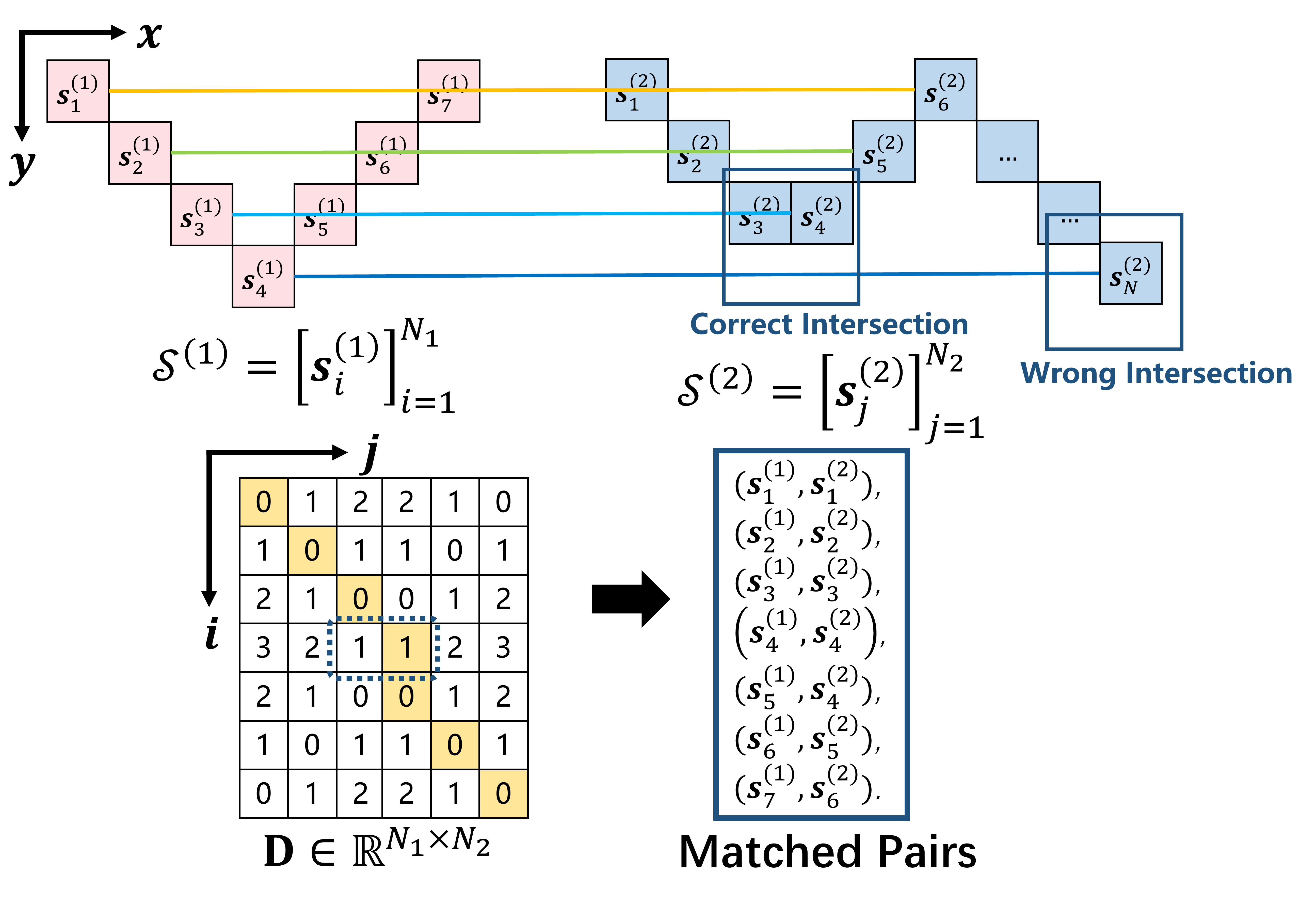}
        \caption{An example of epipolar degeneracy. Missing intersection at correct position is constructed.}
        \label{fig:ECDP_2}
    \end{subfigure}
    \caption{Robust correspondence recovery by ECDP under challenging conditions.(a) Occlusion: ECDP interpolates consistent cross-view point pairs. (b) Epipolar degeneracy: ECDP avoids local mismatches.}
    \label{fig:ECDP}
\end{figure}

After applying the refinement operator $\mathcal{R}$, 
we obtain a unique one-to-one correspondence set $\widehat{\mathcal{M}}
=\bigl\{ (\hat{\mathbf{s}}_{ij}^{(1)}, \mathbf{s}_j^{(2)}) \mid 
\hat{\mathbf{s}}_{ij}^{(1)} = \mathcal{R}(\mathbf{s}_{i-1}^{(1)},\mathbf{s}_i^{(1)},\mathbf{s}_{i+1}^{(1)};\mathbf{D})
\bigr\}_{j=1}^{N_2}$, where $\hat{\mathbf{s}}_{ij}^{(1)}$ denotes the refined point corresponding uniquely to $\mathbf{s}_j^{(2)}$. The final 3D reconstruction is obtained by triangulating each pair 
$(\hat{\mathbf{s}}_{ij}^{(1)}, \mathbf{s}_j^{(2)})$ using the calibrated camera parameters. An example of ECDP is shown in Fig. \ref{fig:ECDP}.

\begin{table}[h]
\centering
\caption{Quantitative comparison on simulation and phantom datasets under different settings. We compare our method against general reconstruction baselines (using GT masks), classical geometry-based methods (using ordered points), and evaluate our end-to-end performance.}
\label{tab:comparison_all}
\resizebox{\textwidth}{!}{
\setlength{\tabcolsep}{3pt} 
\renewcommand{\arraystretch}{1.5} 
\begin{tabular}{l c c cccc c cccc}
\toprule
\multirow{2}{*}{\textbf{Method}} & \multirow{2}{*}{\textbf{Input}} & \multirow{2}{*}{\textbf{Calib.}} & \multicolumn{4}{c}{\textbf{Simulation Dataset (pixel)}} & & \multicolumn{4}{c}{\textbf{Phantom Dataset (mm)}} \\
\cmidrule(lr){4-7} \cmidrule(lr){9-12}
 & & & Acc.$\downarrow$ & Comp.$\downarrow$ & Overall$\downarrow$ & Max Err.$\downarrow$ & & Acc.$\downarrow$ & Comp.$\downarrow$ & Overall$\downarrow$ & Max Err.$\downarrow$ \\
\midrule
\multicolumn{12}{l}{\textit{\textbf{Comparison with General Reconstruction Methods (GT Mask Provided)}}} \\
VGGT~\cite{wang2025vggt}     & Mask & \XSolidBrush & 2.8532 & 2.4845 & 2.6689 & 8.6994 & & 11.9963 & 10.9020 & 11.4492 & 28.9383 \\
Fast3R~\cite{yang2025fast3r}   & Mask & \XSolidBrush & 3.6407 & 3.3810 & 3.5109 & 9.9247 & & 20.7836 & 12.7321 & 16.7579 & 43.1215 \\
MonSter~\cite{cheng2025monster} & Mask & \Checkmark   & 4.1161 & 3.1932 & 3.6547 & 10.7965 & & 4.8306 & 6.1811 & 5.5059 & 14.6172 \\
\textbf{Ours}     & Mask & \Checkmark   & \textbf{0.2363} & \textbf{0.3567} & \textbf{0.2965} & \textbf{2.7738} & & \textbf{0.4475} & \textbf{0.4959} & \textbf{0.4717} & \textbf{1.6256} \\

\midrule
\multicolumn{12}{l}{\textit{\textbf{Comparison with Classical Geometry Method (Ordered Points Provided)}}} \\
Baert \textit{et al.}~\cite{baert2003three} & Points & \Checkmark & 0.8843 & 0.5570 & 0.7207 & 5.2142 & & 0.4051 & 0.3096 & 0.3573 & 1.9706 \\
Hoffmann \textit{et al.}~\cite{hoffmann2015electrophysiology} & Points & \Checkmark & 0.1023 & 0.1195 & 0.1109 & 1.3444 & & 0.4632 & 0.3315 & 0.3974 & 1.3208 \\
\textbf{Ours}                   & Points & \Checkmark & \textbf{0.0682} & \textbf{0.0636} & \textbf{0.0659} & \textbf{0.4378} &  & \textbf{0.3880} & \textbf{0.2724} & \textbf{0.3302} & \textbf{0.9457} \\

\midrule
\multicolumn{12}{l}{\textit{\textbf{End-to-End Full Pipeline Performance (Raw Images Input)}}} \\
\textbf{Ours}     & Image & \Checkmark   & 0.3118 & 0.2512 & 0.2815 & 2.4049 & & 0.5965 & 0.6188 & 0.6076 & 2.0604 \\
\bottomrule
\end{tabular}
}
\end{table}

\section{Experiment}
\label{sec: experiments}

We evaluate ACCURATE on both simulated and real phantom data to assess reconstruction accuracy and generalization across diverse continuum shapes. For simulated data, we synthetically generate long slender continuum structures with highly deformable and complex geometries, where strong non-planar bending naturally leads to self-occlusion in projection. Using a virtual calibrated dual-view imaging setup built in Blender, we render projection images and exact 3D ground truth. For real phantom data, the fluoroscopic data of navigating clinically used guidewires (0.035" and 0.014" in diameter) in a 1:1 full-body vascular phantom, simulating cardiac and cerebrovascular intervention, is acquired using a GE INNOVA angiographic system. Three-dimensional reconstruction is performed on an Advantage Workstation (AW 4.6) using Innova3DXR software. Performance is assessed using Accuracy (Acc.), Completeness (Comp.), Overall Chamfer Distance(Overall), and Maximum Error(Max Err.). Table~\ref{tab:comparison_all} summarizes results across different experimental settings.

To provide a comprehensive evaluation, we select representative benchmark methods such as VGGT~\cite{wang2025vggt},Fast3R~\cite{yang2025fast3r}, and MonSter~\cite{cheng2025monster}, representing the current state of the art in general 3D reconstruction. Classical geometry-based approaches such as Baert \textit{et al.}~\cite{baert2003three} reconstruct correspondences by computing intersections between epipolar lines and fitted curves, while Hoffmann \textit{et al.}~\cite{hoffmann2015electrophysiology} also employ dynamic programming to construct correspondences of intersections. These methods are explicitly tailored to continuum-shaped objects and serve as strong geometry-based baselines.

\begin{figure}[h]
    \centering
    \begin{minipage}[b]{0.433\linewidth}
        \centering
        \includegraphics[width=\linewidth]{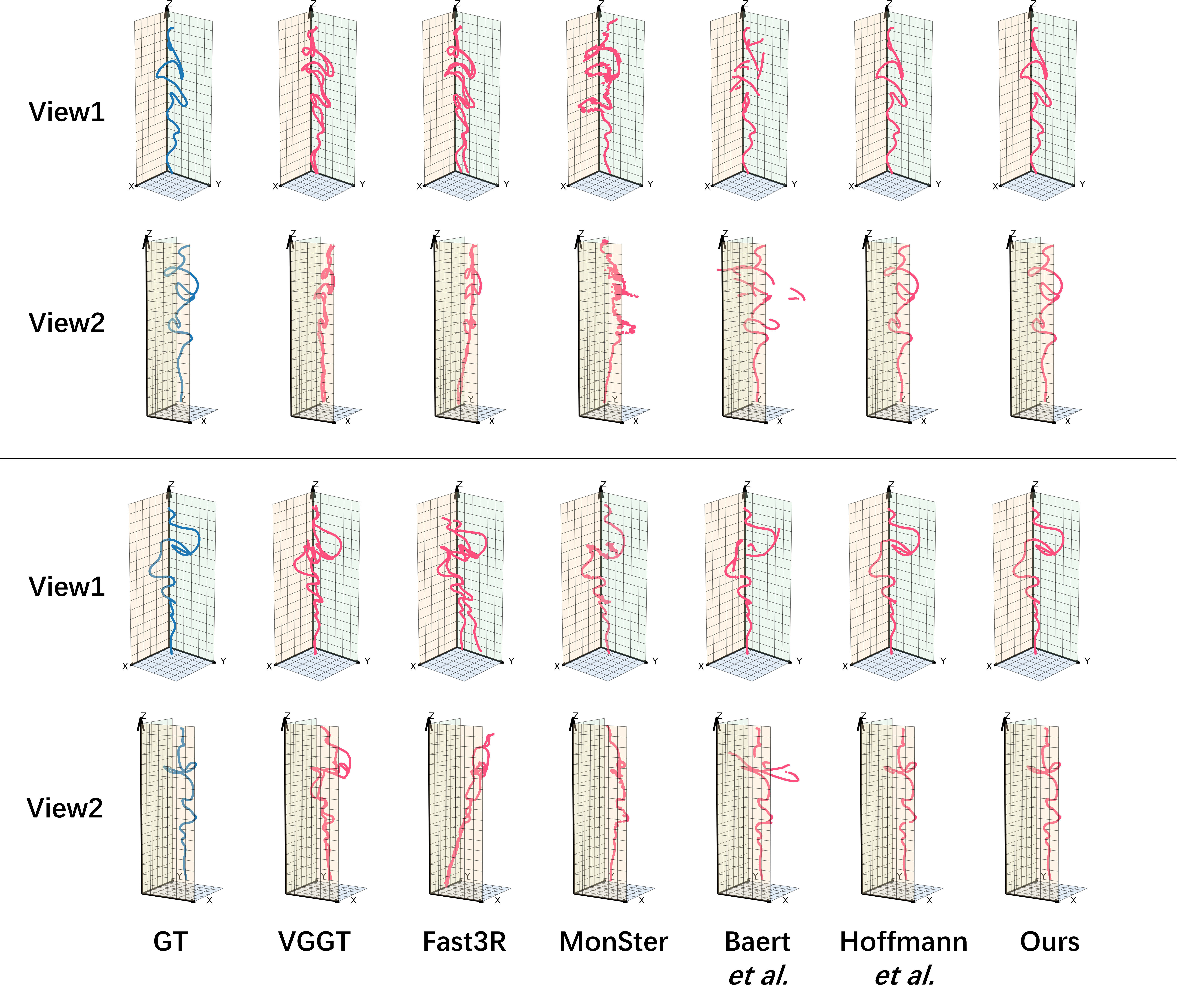}
        \caption{GT and reconstruction results on simulation dataset.}
        \label{fig:simulation_result}
    \end{minipage}
    \hspace{0.02\linewidth} 
    \begin{minipage}[b]{0.527\linewidth}
        \centering
        \includegraphics[width=\linewidth]{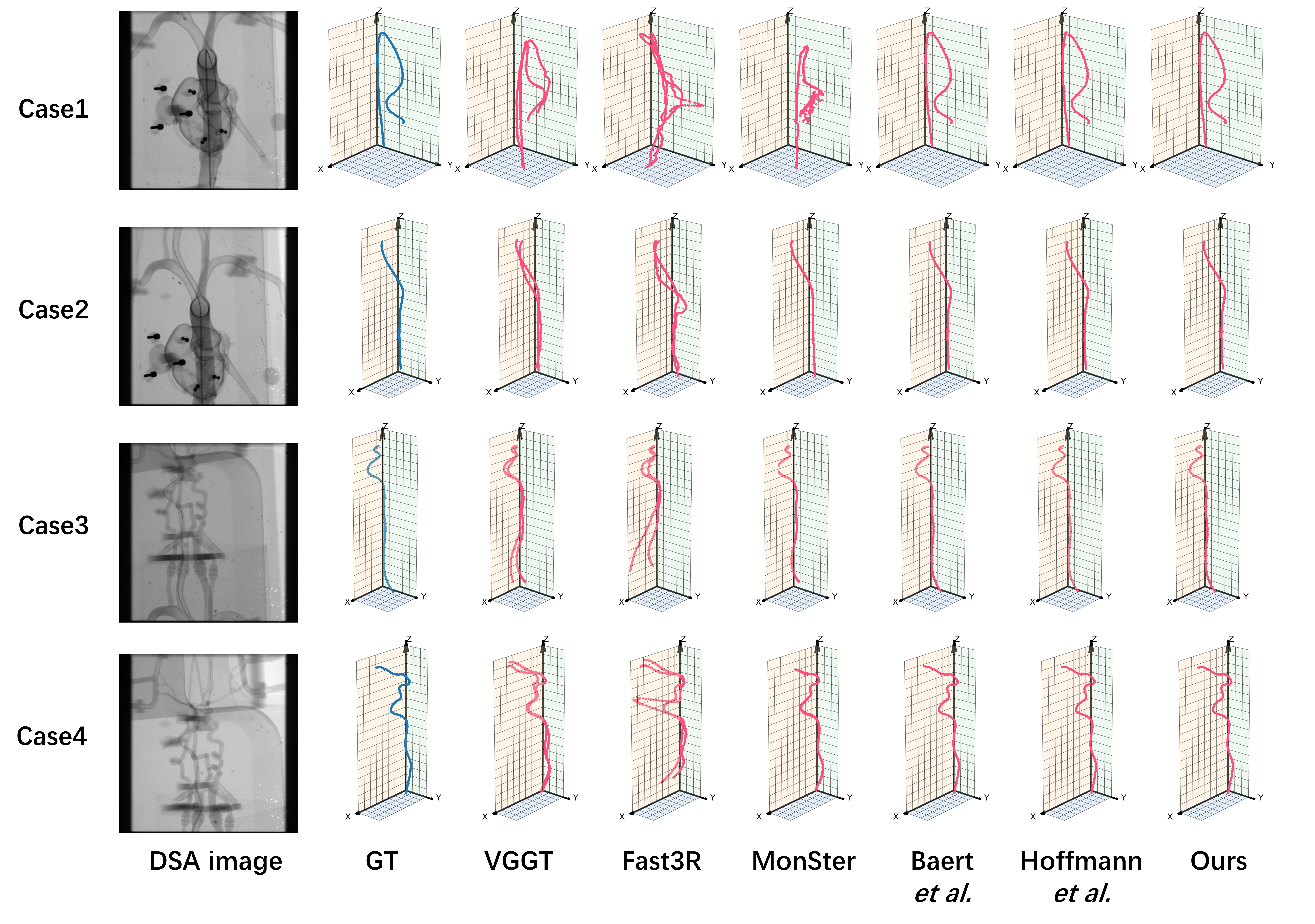}
        \caption{Image, GT and reconstruction results on real phantom dataset.}
        \label{fig:phantom_result}
    \end{minipage}
\end{figure}

For VGGT~\cite{wang2025vggt}, Fast3R~\cite{yang2025fast3r}, and MonSter~\cite{cheng2025monster}, direct application leads to full-scene reconstruction thus two-dimensional GT masks are used to isolate the target structures. To ensure a fair comparison regarding geometric reconstruction capability, our method is also evaluated using GT masks. As shown in Table~\ref{tab:comparison_all}, Fig.~\ref{fig:simulation_result} and Fig.~\ref{fig:phantom_result}, these methods yield relatively large errors, highlighting the necessity of designing algorithms tailored to long slender bodies.

To validate this motivation, classical geometry-based reconstruction methods such as Baert \textit{et al.}~\cite{baert2003three}, which is applied in reconstruction step of Guide3D~\cite{jianu2024guide3d}, and Hoffmann \textit{et al.}~\cite{hoffmann2015electrophysiology} are compared. Under the same setting, ACCURATE demonstrates improved accuracy on geometrically complex structures, indicating better generalizability. In contrast, strict intersection-based methods may encounter incorrect intersections for complex curves, whereas our approach leverages global distance information for more robust matching. On the real phantom dataset containing relatively simple shapes, despite small calibration errors, our method still achieves consistent accuracy gains.

Furthermore, we evaluate a full-pipeline scenario using images and camera parameters, as shown in Fig.~\ref{fig:simulation_result} and Fig.~\ref{fig:phantom_result}, with quantitative results reported in Table~\ref{tab:comparison_all}. Compared with providing ordered points, increased error indicates that segmentation accuracy becomes the performance bottleneck.

ACCURATE achieves good performance across all evaluation metrics with different input. Experiment results indicate that fully exploiting camera geometry and designing reconstruction algorithms specifically for long slender bodies are crucial for achieving accurate and robust 3D reconstruction.

\section{Conclusion}

In this work, we presented ACCURATE, a biplanar imaging-based 3D reconstruction framework for arbitrary-shaped long slender continuum bodies that integrates topology-aware perception with explicit geometric enforcement. ACCURATE combines segmentation-driven structural understanding with globally constrained epipolar optimization. By integrating geometry-consistent topology traversal with a dynamic programming-based solver, this framework achieves robust reconstruction under large deformations, occlusions, discretization and degenerate epipolar lines. Experiments on both simulated and real phantom data demonstrate that ACCURATE achieves substantially improved reconstruction accuracy, reaching sub-millimeter reconstruction errors, leading to reliable downstream mechanical simulation and intervention guidance. Furthermore, publicly released calibrated datasets and codebase establish a standardized benchmark for future research on slender continuum reconstruction.

\section*{Acknowledgements}

This work was supported by National Natural Science Foundation of China (52405282).

\bibliographystyle{splncs04}
\bibliography{myrefs}

@article{diezinger20223d,
  title={3D curvature-based tip load estimation for continuum robots},
  author={Diezinger, Matyas A and Tamadazte, Brahim and Laurent, Guillaume J},
  journal={IEEE Robotics and Automation Letters},
  volume={7},
  number={4},
  pages={10526--10533},
  year={2022},
  publisher={IEEE}
}

@article{sahu2021shape,
  title={Shape reconstruction processes for interventional application devices: State of the art, progress, and future directions},
  author={Sahu, Sujit Kumar and Sozer, Canberk and Rosa, Benoit and Tamadon, Izadyar and Renaud, Pierre and Menciassi, Arianna},
  journal={Frontiers in Robotics and AI},
  volume={8},
  pages={758411},
  year={2021},
  publisher={Frontiers Media SA}
}

@article{ourak2021fusion,
  title={Fusion of biplane fluoroscopy with fiber Bragg grating for 3D catheter shape reconstruction},
  author={Ourak, Mouloud and De Buck, Stijn and Ha, Xuan Thao and Al-Ahmad, Omar and Bamps, Kobe and Ector, Joris and Vander Poorten, Emmanuel},
  journal={IEEE Robotics and Automation Letters},
  volume={6},
  number={4},
  pages={6505--6512},
  year={2021},
  publisher={IEEE}
}

@article{lu2023robust,
  title={A robust graph-based framework for 3-d shape reconstruction of flexible medical instruments using multi-core fbgs},
  author={Lu, Yiang and Chen, Wei and Li, Bin and Lu, Bo and Zhou, Jianshu and Chen, Zhi and Liu, Yun-Hui},
  journal={IEEE Transactions on Medical Robotics and Bionics},
  volume={5},
  number={3},
  pages={472--485},
  year={2023},
  publisher={IEEE}
}

@article{ha2023sensor,
  title={Sensor fusion for shape reconstruction using electromagnetic tracking sensors and multi-core optical fiber},
  author={Ha, Xuan Thao and Wu, Di and Ourak, Mouloud and Borghesan, Gianni and Menciassi, Arianna and Vander Poorten, Emmanuel},
  journal={IEEE Robotics and Automation Letters},
  volume={8},
  number={7},
  pages={4076--4083},
  year={2023},
  publisher={IEEE}
}

@book{harrigan2012handbook,
  title={Handbook of cerebrovascular disease and neurointerventional technique},
  author={Harrigan, Mark R and Deveikis, John P},
  volume={1},
  year={2012},
  publisher={Springer Science \& Business Media}
}

@article{baert2003three,
  title={Three-dimensional guide-wire reconstruction from biplane image sequences for integrated display in 3-D vasculature},
  author={Baert, Shirley AM and Van de Kraats, Everine B and Van Walsum, Theo and Viergever, Max A and Niessen, Wiro J},
  journal={IEEE transactions on medical imaging},
  volume={22},
  number={10},
  pages={1252--1258},
  year={2003},
  publisher={IEEE}
}

@inproceedings{hoffmann2012semi,
  title={Semi-automatic catheter reconstruction from two views},
  author={Hoffmann, Matthias and Brost, Alexander and Jakob, Carolin and Bourier, Felix and Koch, Martin and Kurzidim, Klaus and Hornegger, Joachim and Strobel, Norbert},
  booktitle={International Conference on Medical Image Computing and Computer-Assisted Intervention},
  pages={584--591},
  year={2012},
  organization={Springer}
}

@article{hoffmann2015electrophysiology,
  title={Electrophysiology catheter detection and reconstruction from two views in fluoroscopic images},
  author={Hoffmann, Matthias and Brost, Alexander and Koch, Martin and Bourier, Felix and Maier, Andreas and Kurzidim, Klaus and Strobel, Norbert and Hornegger, Joachim},
  journal={IEEE transactions on medical imaging},
  volume={35},
  number={2},
  pages={567--579},
  year={2015},
  publisher={IEEE}
}

@article{miyato2023gta,
  title={Gta: A geometry-aware attention mechanism for multi-view transformers},
  author={Miyato, Takeru and Jaeger, Bernhard and Welling, Max and Geiger, Andreas},
  journal={arXiv preprint arXiv:2310.10375},
  year={2023}
}

@inproceedings{cheng2025monster,
  title={Monster: Marry monodepth to stereo unleashes power},
  author={Cheng, Junda and Liu, Longliang and Xu, Gangwei and Wang, Xianqi and Zhang, Zhaoxing and Deng, Yong and Zang, Jinliang and Chen, Yurui and Cai, Zhipeng and Yang, Xin},
  booktitle={Proceedings of the Computer Vision and Pattern Recognition Conference},
  pages={6273--6282},
  year={2025}
}

@inproceedings{jianu2024guide3d,
  title={Guide3D: A Bi-planar X-ray Dataset for Guidewire Segmentation and 3D Reconstruction},
  author={Jianu, Tudor and Huang, Baoru and Nguyen, Hoan and Bhattarai, Binod and Do, Tuong and Tjiputra, Erman and Tran, Quang and Berthet-Rayne, Pierre and Le, Ngan and Fichera, Sebastiano and others},
  booktitle={Proceedings of the Asian Conference on Computer Vision},
  pages={1549--1565},
  year={2024}
}

@inproceedings{wang2025vggt,
  title={Vggt: Visual geometry grounded transformer},
  author={Wang, Jianyuan and Chen, Minghao and Karaev, Nikita and Vedaldi, Andrea and Rupprecht, Christian and Novotny, David},
  booktitle={Proceedings of the Computer Vision and Pattern Recognition Conference},
  pages={5294--5306},
  year={2025}
}

@inproceedings{yang2025fast3r,
  title={Fast3r: Towards 3d reconstruction of 1000+ images in one forward pass},
  author={Yang, Jianing and Sax, Alexander and Liang, Kevin J and Henaff, Mikael and Tang, Hao and Cao, Ang and Chai, Joyce and Meier, Franziska and Feiszli, Matt},
  booktitle={Proceedings of the Computer Vision and Pattern Recognition Conference},
  pages={21924--21935},
  year={2025}
}

@article{zhang1998determining,
  title={Determining the epipolar geometry and its uncertainty: A review},
  author={Zhang, Zhengyou},
  journal={International journal of computer vision},
  volume={27},
  number={2},
  pages={161--195},
  year={1998},
  publisher={Springer}
}

@inproceedings{yao2018mvsnet,
  title={Mvsnet: Depth inference for unstructured multi-view stereo},
  author={Yao, Yao and Luo, Zixin and Li, Shiwei and Fang, Tian and Quan, Long},
  booktitle={Proceedings of the European conference on computer vision (ECCV)},
  pages={767--783},
  year={2018}
}

@article{isensee2021nnu,
  title={nnU-Net: a self-configuring method for deep learning-based biomedical image segmentation},
  author={Isensee, Fabian and Jaeger, Paul F and Kohl, Simon AA and Petersen, Jens and Maier-Hein, Klaus H},
  journal={Nature methods},
  volume={18},
  number={2},
  pages={203--211},
  year={2021},
  publisher={Nature Publishing Group}
}

@inproceedings{shit2021cldice,
  title={clDice-a novel topology-preserving loss function for tubular structure segmentation},
  author={Shit, Suprosanna and Paetzold, Johannes C and Sekuboyina, Anjany and Ezhov, Ivan and Unger, Alexander and Zhylka, Andrey and Pluim, Josien PW and Bauer, Ulrich and Menze, Bjoern H},
  booktitle={Proceedings of the IEEE/CVF conference on computer vision and pattern recognition},
  pages={16560--16569},
  year={2021}
}

@inproceedings{kirchhoff2024skeleton,
  title={Skeleton recall loss for connectivity conserving and resource efficient segmentation of thin tubular structures},
  author={Kirchhoff, Yannick and Rokuss, Maximilian R and Roy, Saikat and Kovacs, Balint and Ulrich, Constantin and Wald, Tassilo and Zenk, Maximilian and Vollmuth, Philipp and Kleesiek, Jens and Isensee, Fabian and others},
  booktitle={European Conference on Computer Vision},
  pages={218--234},
  year={2024},
  organization={Springer}
}

@article{lu2021toward,
  title={Toward image-guided automated suture grasping under complex environments: A learning-enabled and optimization-based holistic framework},
  author={Lu, Bo and Li, Bin and Chen, Wei and Jin, Yueming and Zhao, Zixu and Dou, Qi and Heng, Pheng-Ann and Liu, Yunhui},
  journal={IEEE Transactions on Automation Science and Engineering},
  volume={19},
  number={4},
  pages={3794--3808},
  year={2021},
  publisher={IEEE}
}

@inproceedings{hoffmann2013reconstruction,
  title={Reconstruction method for curvilinear structures from two views},
  author={Hoffmann, Matthias and Brost, Alexander and Jakob, Carolin and Koch, Martin and Bourier, Felix and Kurzidim, Klaus and Hornegger, Joachim and Strobel, Norbert},
  booktitle={Medical Imaging 2013: Image-Guided Procedures, Robotic Interventions, and Modeling},
  volume={8671},
  pages={630--637},
  year={2013},
  organization={SPIE}
}

@article{altingovde20223d,
  title={3D reconstruction of curvilinear structures with stereo matching deep convolutional neural networks},
  author={Alting{\"o}vde, Okan and Mishchuk, Anastasiia and Ganeeva, Gulnaz and Oveisi, Emad and Hebert, Cecile and Fua, Pascal},
  journal={Ultramicroscopy},
  volume={234},
  pages={113460},
  year={2022},
  publisher={Elsevier}
}

@inproceedings{ambrosini20153d,
  title={3D catheter tip tracking in 2D X-ray image sequences using a hidden Markov model and 3D rotational angiography},
  author={Ambrosini, Pierre and Smal, Ihor and Ruijters, Daniel and Niessen, Wiro J and Moelker, Adriaan and van Walsum, Theo},
  booktitle={Workshop on Augmented Environments for Computer-Assisted Interventions},
  pages={38--49},
  year={2015},
  organization={Springer}
}

@article{baur2016automatic,
  title={Automatic 3D reconstruction of electrophysiology catheters from two-view monoplane C-arm image sequences},
  author={Baur, Christoph and Milletari, Fausto and Belagiannis, Vasileios and Navab, Nassir and Fallavollita, Pascal},
  journal={International journal of computer assisted radiology and surgery},
  volume={11},
  number={7},
  pages={1319--1328},
  year={2016},
  publisher={Springer}
}

@inproceedings{robertshaw2025world,
  title={World Model for AI Autonomous Navigation in Mechanical Thrombectomy},
  author={Robertshaw, Harry and Wu, Han-Ru and Granados, Alejandro and Booth, Thomas C},
  booktitle={International Conference on Medical Image Computing and Computer-Assisted Intervention},
  pages={680--690},
  year={2025},
  organization={Springer}
}

@inproceedings{cafaro2024two,
  title={Two projections suffice for cerebral vascular reconstruction},
  author={Cafaro, Alexandre and Dorent, Reuben and Haouchine, Nazim and Lepetit, Vincent and Paragios, Nikos and Wells III, William M and Frisken, Sarah},
  booktitle={International Conference on Medical Image Computing and Computer-Assisted Intervention},
  pages={722--731},
  year={2024},
  organization={Springer}
}
\end{document}